\documentclass[10pt,twocolumn,letterpaper]{article}

\usepackage{iccv}
\usepackage{times}
\usepackage{epsfig}
\usepackage{graphicx}
\usepackage{amsmath}
\usepackage{amssymb}
\usepackage{algorithm}
\usepackage{algorithmic}
\usepackage{pifont}
\usepackage[misc]{ifsym}
\usepackage[accsupp]{axessibility}  

\usepackage{booktabs}
\usepackage{lineno}
\modulolinenumbers[5]
\usepackage{multirow}
\usepackage{array}
\usepackage{float}

\newcolumntype{C}[1]{>{\centering\let\newline\\\arraybackslash}m{#1}}

\usepackage[pagebackref=true,breaklinks=true,letterpaper=true,colorlinks,bookmarks=false]{hyperref}

\iccvfinalcopy 

\def\iccvPaperID{8161} 

\ificcvfinal\fi

\begin{document}

\title{Multi-Anchor Active Domain Adaptation for Semantic Segmentation}

\author{Munan Ning\thanks{Contributed equally.}, Donghuan Lu\footnotemark[1], Dong Wei\thanks{Correspondence: donwei@tencent.com}, Cheng Bian, Chenglang Yuan, \\
Shuang Yu, Kai Ma, Yefeng Zheng\\
Tencent Jarvis Lab, Shenzhen, China\\

}

\maketitle
\ificcvfinal\thispagestyle{empty}\fi

\begin{figure*}[!ht]
	\centering
	\includegraphics[width=1.7\columnwidth]{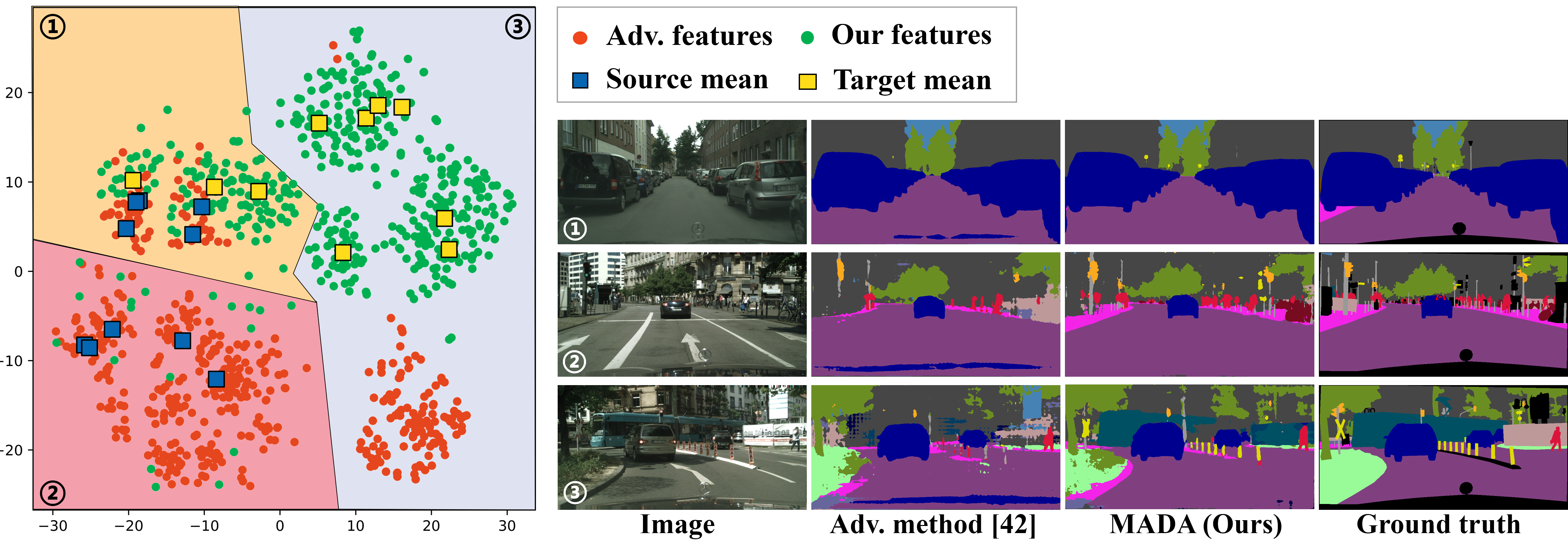}
	\caption{Visualization (t-SNE~\cite{hinton2002stochastic}) of the target-domain distribution distortion problem in UDA (left). The blue and yellow squares are the average latent representation of different category samples from the target-domain extracted by two networks trained with the source- and target-domain data, respectively, where we can observe little overlap (region \ding{172}) along with large discrepancy (regions \ding{173} and  \ding{174}) between the two distributions.
    The red dots denote the adapted target-domain features by a typical adversarial (adv.) training based UDA method \cite{tsai2018learning}. A general alignment to the source-domain distribution can be observed in region \ding{172},  while an obvious distortion of the target-domain distribution is displayed in region \ding{173} and \ding{174}, yielding unsatisfactory performance as presented on the right figures. By adopting active learning for the domain adaptation, such distortion is effectively alleviated, as demonstrated by the correctly distributed green dots.}
	\label{fig:cluster_visual_1}
	\vspace{-0.3cm}
\end{figure*}

\begin{abstract}
Unsupervised domain adaption has proven to be an effective approach for alleviating the intensive workload of manual annotation by aligning the synthetic source-domain data and the real-world target-domain samples. Unfortunately, mapping the target-domain distribution to the source-domain unconditionally may distort the essential structural information of the target-domain data. To this end, we firstly propose to introduce a novel multi-anchor based active learning strategy to assist domain adaptation regarding the semantic segmentation task. By innovatively adopting multiple anchors instead of a single centroid, the source domain can be better characterized as a multimodal distribution, thus more representative and complimentary samples are selected from the target domain. With little workload to manually annotate these active samples, the distortion of the target-domain distribution can be effectively alleviated, resulting in a large performance gain. The multi-anchor strategy is additionally employed to model the target-distribution. By regularizing the latent representation of the target samples compact around multiple anchors through a novel soft alignment loss, more precise segmentation can be achieved. Extensive experiments are conducted on public datasets to demonstrate that the proposed approach outperforms state-of-the-art methods significantly, along with thorough ablation study to verify the effectiveness of each component. The code will be released soon at https://github.com/munanning/MADA.

\end{abstract}

\section{Introduction}



Semantic segmentation has always been a fundamental task in computer vision.
Benefiting from the rapid development of deep learning, many advanced segmentation methods have been proposed and achieved great breakthroughs with high accuracies for various tasks, such as autonomous driving~\cite{geiger2012we}, scene parsing~\cite{cordts2016cityscapes,Piao_2019_DMRA}, object detection~\cite{Ji_2021_DCF,Zhang_2019_MoLF} and human-computer interaction~\cite{oberweger2015hands}.
However, the requirement of large amount of data with accurate pixel-wise annotation limits their usage in many practical applications, \textit{e.g.}, medical image segmentation~\cite{Ji_2021_MRNet,ning2020macro,ning2021ensembled,ning2021new,ma2021abdomenct} and auto-driving tasks~\cite{choi2020cars}.

To avoid the intensive workload of manual annotation, a lot of efforts have been made on unsupervised domain adaptation (UDA)~\cite{chen2017no,hoffman2018cycada,hoffman2016fcns,tsai2018learning}, which aims at aligning the target-domain distribution towards the source-domain distribution, so that networks trained with the supervision of only the synthetic source data can be applied to the real-world target data. However, forcing the target-domain features to fit the source-domain distribution may destroy the latent structural pattern of the target domain, resulting in inferior performance. 
As illustrated by the t-SNE~\cite{hinton2002stochastic} visualization in Fig.~\ref{fig:cluster_visual_1}, the distributions of the source and target domains present both overlap (region \ding{172}) and obvious discrepancies (regions \ding{173} and  \ding{174}). When the adapted target-domain features (red dots) obtained with a typical UDA method based on adversarial training \cite{tsai2018learning}---despite generally aligned with the source domain distributions (blue squares)---show a clear distortion of the target-domain distribution in region \ding{173}, the adapted network presents unsatisfactory performance. Worse segmentation can be observed in region \ding{174} when some specific targets are aligned neither with the source domain nor the target domain.
A promising strategy to efficiently prevent such distortion of the target-domain distribution with minimal annotation workload is active learning (AL) \cite{settles2009active}. By introducing little extra manual annotation for a few selected samples from the target domain, the performance can be significantly boosted regarding the classification and the detection tasks \cite{su2020active}. However, the sample selection methods in all previous active learning studies \cite{su2020active} assumed a unimodal source-domain distribution and neglected the potential multimodal distribution, resulting in sub-optimal active samples and inferior performance, as demonstrated in Table~\ref{table:selection}. 


To address the above issues, we firstly propose to adopt the active learning strategy to assist domain adaptation (DA) regarding the semantic segmentation task, so that the essential structural pattern of the target domain can be maintained with minimal manual annotation workload. In addition, a multi-anchor strategy is proposed to better characterize the source-domain features as well as the target-domain features. Specifically, the proposed Multi-anchor Active Domain Adaptation (MADA) framework consists of two stages. In the first stage, with the network pretrained in an adversarial UDA \cite{tsai2018learning} manner, a multi-anchor based active sample selection strategy is proposed to identify the most complementary and representative samples for manual annotation by exploiting the feature distributions across the target and source domains. 
Then in the second stage, the segmentation network is fine-tuned in a semi-supervised learning manner. The annotations of the source samples and the few selected target samples are used for supervision, while all the available image information is additionally applied for optimization with a pseudo label loss and the proposed multi-anchor soft-alignment loss. In summary, our paper makes the following contributions:

\begin{itemize}

\item To the best of our knowledge, our work is the first study to adopt active learning to assist the domain adaptation regarding the semantic segmentation tasks. With little manual annotation workload of few target-domain samples,
the distortion of the target-domain feature distribution can be effectively prevented and superior segmentation performance can be achieved.

\item Assuming a multimodal distribution in practical situations, we propose to adopt multiple anchors obtained via clustering-based method to characterize the feature distribution of the source-domain, so that the representative target-domain samples which are the most complimentary to the source-domain can be selected.

\item The multi-anchor strategy is used further to model the target-domain feature distribution. With the proposed multi-anchor soft-alignment loss, we show that explicitly pushing the features of the target samples towards multiple anchors leads to better latent representation, thus notably improve the segmentation performance.

\item We conduct extensive experiments to demonstrate the superiority of the proposed MADA framework, along with thorough ablation studies to evaluate the effectiveness of the multi-anchor strategy on modeling the feature distribution.

\end{itemize}

\section{Related Work}
\subsection{Unsupervised Domain Adaptation}
Unsupervised domain adaptation (UDA) has been proposed for years, aiming to address the domain shift problem in a wide variety of computer vision tasks including classification~\cite{glorot2011domain}, detection~\cite{chen2018domain}, and segmentation~\cite{tsai2018learning}.
Recent UDA methods can be roughly divided into two groups: maximum mean discrepancy (MMD) based and adversarial learning based.
The MMD kernel was first introduced in~\cite{long2015learning}, which measured the discrepancy of features from different domains quantitatively.
Subsequent studies proposed several improved MMD kernels for more accurate measurement of the domain discrepancy, including MK-MMD~\cite{long2015learning}, JMMD~\cite{long2017deep}, CMD~\cite{zellinger2017central} and CORAL~\cite{sun2016deep}. 
Minimization of the discrepancy yielded by these kernels forced features from different domains to align with each other, thus addressing the domain shift problem.
However, it is impractical to directly adopt the MMD-based methods in segmentation tasks, because these methods required complex computation in the high-dimension feature space.

In contrast, adversarial learning based methods are preferred for UDA of segmentation tasks, where the two domain distributions are drawn together via a domain discriminator.
The classical appearance matching method CycleGAN~\cite{zhu2017unpaired} constructed two adversarial subnets to translate unpaired source and target images.
BDL~\cite{li2019bidirectional} leveraged label consistency to improve the UDA performance.
DISE~\cite{chang2019all} proposed a disentangled representation learning architecture~\cite{huang2018multimodal} to preserve structural information during image translation.
Feature aligning methods such as CLAN~\cite{luo2019taking} and CAG~\cite{zhang2019category} utilized category-based distribution alignment to adapt the source and target domains in the feature and output spaces.
AdvEnt~\cite{vu2019advent} designed a novel loss function to maximize the prediction certainty in the target domain to boost the UDA performance.

Despite the encouraging progress, UDA methods 
unconditionally force the distributions of the two domains to be similar, which may distort the underlying latent distribution of the target domain if it presents intrinsic difference from that of the source domain.
A promising strategy to prevent such distortion with minimal annotation workload is active learning (AL) \cite{settles2009active}, which we adopt in this work.

\begin{figure*}[!htp]
	\centering
	\includegraphics[width=2.0\columnwidth]{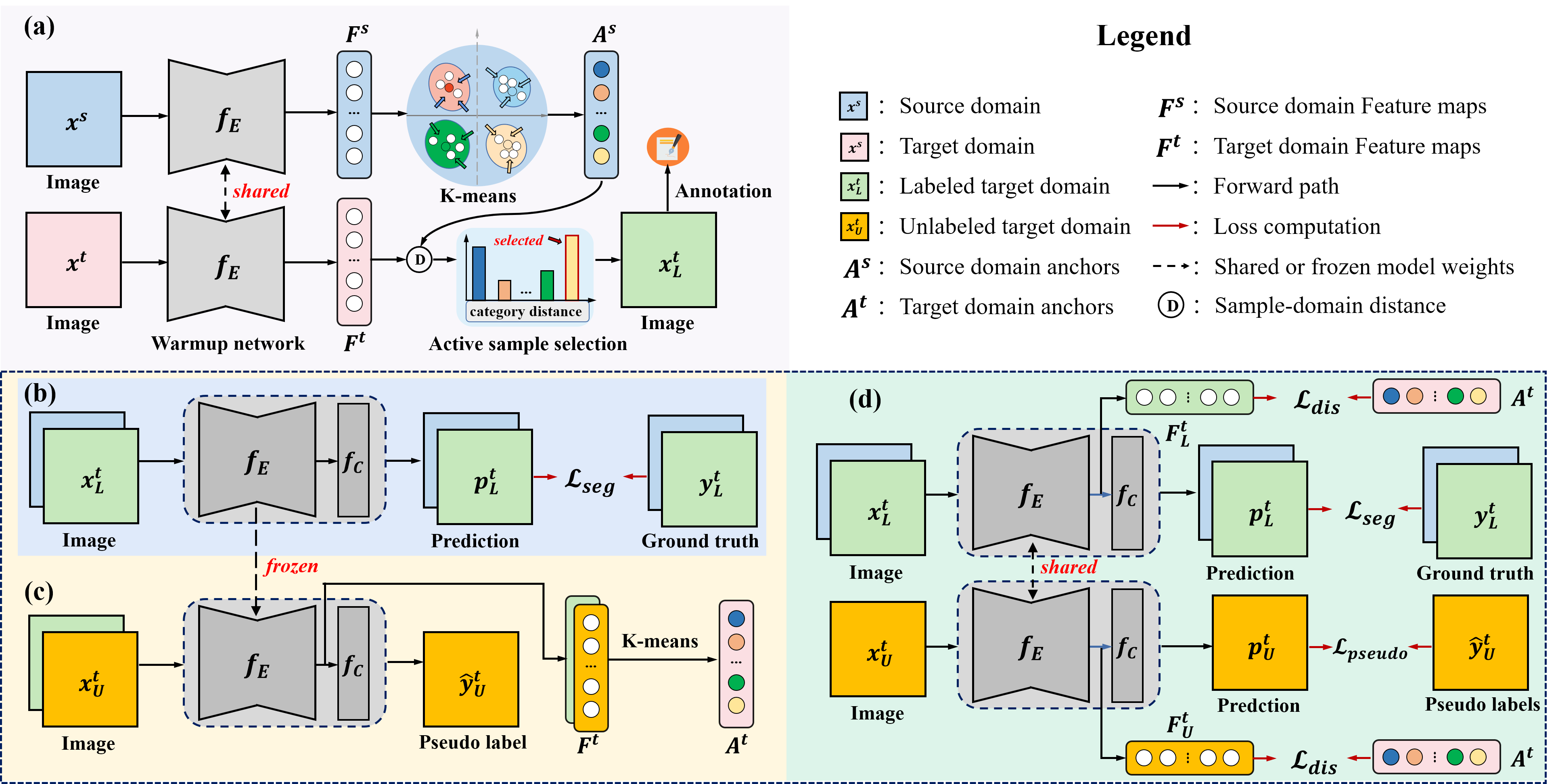}
	\caption{Overview of the proposed MADA framework.}
	\label{fig:framework}
	\vspace{-0.3cm}
\end{figure*}

\subsection{Active Learning and Domain Adaptation}
AL aims at optimal performance at a low annotation cost, by actively selecting the few samples that are most helpful to performance improvement, if labeled~\cite{cohn1996active}.
Over the past decade, several sample selection strategies have been proposed for AL,
including uncertainty-based~\cite{lewis1994heterogeneous,scheffer2001active}, diversity-based~\cite{dutt2016active,hoi2009semisupervised}, representativeness-based~\cite{huang2010active,dasgupta2008hierarchical,nguyen2004active}, and
expected model change based~\cite{freytag2014selecting,kading2015active,vezhnevets2012weakly}. 
These strategies have been successfully applied to various computer vision tasks, such as image classification~\cite{qi2008two}, object detection~\cite{kao2018localization,Ji_2020_CoNet,Zhang_2020_LFNet}, and image segmentation~\cite{sun2015active}.
In this work, we argue that it is beneficial to introduce AL to the DA problem, to avoid distortion of the target-domain distribution.
First, AL only entails minimal annotation cost, which is acceptable in many scenarios considering the potential performance gain.
Second, with a proper sample selection strategy, AL can identify the samples most representative of the exclusive components in the target-domain distribution for annotation.
Hence, how to select the AL samples becomes a critical issue.

As far as the authors are aware of, only few studies attempted applying AL to DA problems.
An early work by Chattopadhyay \emph{et al.}~\cite{chattopadhyay2013joint} 
proposed to use the MMD distance between the source and target domains for active sample selection during the DA process.
However, it is practically prohibitive to apply MMD distances for segmentation DA problems, as mentioned earlier.
More recently, Huang \emph{et al.}~\cite{huang2018cost} proposed to fine-tune pre-trained models for classification tasks and involved additional active sample selection in every iteration.
In contrast, our framework takes a step forward to make dense
predictions for segmentation tasks, and simplifies the active
learning process to a one-time sample selection.
Being closely related to our work, Active Adversarial Domain Adaptation (AADA)~\cite{su2020active} proposed AL for DA with the adversarial learning~\cite{ganin2016domain} strategy, where representative samples were selected by jointly considering diversity and uncertainty criteria.
In this work, by modeling both the source and target distributions as multimodal (in contrast to the implicit unimodal assumption in previous works such as AADA), our method captures more comprehensive information from both domains and can achieve substantial performance improvement (experimentally validated in Section~\ref{sec:sample_selection}).

\section{Method}
The proposed method consists of two main stages: active target sample selection based on multiple anchors of the source domain (Fig. \ref{fig:framework}(a)), and semi-supervised domain adaptation enhanced by a novel multi-anchor soft alignment loss (Figs. \ref{fig:framework}(b), \ref{fig:framework}(c) and \ref{fig:framework}(d)).
Below we first formally define our problem setting, then elaborate the two stages.

\subsection{Problem Setting}
The goal of semantic segmentation is to train a model $\mathbf{M}$ to map a sample $x$ in the image space $X$ to a prediction $y$ in the label space $Y$, where $x \in \mathbb{R}^{H \times W \times 3}$ with $H$ denoting the height, $W$ for the width, and 3 for the color channels, and $y \in\{0,1\}^{H \times W \times C}$ with $C$ denoting the number of segmentation categories.
For DA, there are $N_{s}$ image-label pairs $X^s=\left\{\left(x^{s}, y^{s}\right)\right\}$ in the source domain, and $N_{t}$ unlabeled images $X^t=\left\{x^{t}\right\}$ in the target domain.
For AL, $N_{a}$ active samples are selected in the target domain for annotation, where $N_{a} \ll N_t$, so that the target-domain data consist of $N_{a}$ image-label pairs $X^t_L=\left\{\left(x^{t}_{L}, y^{t}_{L}\right)\right\}$ and $N_{t}-N_{a}$ unlabeled images $X^t_U=\left\{x^{t}_{U}\right\}$.
Given the scenario, the target of this work is to optimize the segmentation performance of $\mathbf{M}$ in the target domain while keeping $N_a$ small.


\subsection{Multi-anchor based Active Sample Selection}
\noindent
\textbf{Multiple Anchoring Mechanism.}
In this work, we propose an efficient anchoring mechanism to model the domain distributions, and close the gap between network predictions and the anchors by forming compact clusters around the anchors.
Previously, CAG~\cite{zhang2019category} averaged all image-level features of the source domain to obtain a centroid representing the entire domain, 
which implicity assumed a unimodal distribution.
In practice, however, the distribution of a domain may actually comprise more than a single mode \cite{cui2020unified}.
Although different images may contain the same categories of objects (\textit{e.g.}, road, car, human, and vegetable), they can be classified into various scenes (\textit{e.g.}, highway, uptown, and suburb) based on their overall representative distributions. By concatenating the features of different categories into an image-level `connected' vector, we perform clustering on them to estimate scene-specific representative distributions with cluster centers, denoted as ‘anchors’. We then measure the distance between each target sample and its nearest source anchor, and select the furthest samples.
Below we first elaborate our multiple anchoring mechanism (demonstrated with the source domain), then describe how to use it for active target sample selection.

As a warm-up, we first employ the common adversarial training~\cite{tsai2018learning} strategy to narrow the gap between the source and target domains.
After that, we freeze the feature encoder $f_E$ and calculate the feature map $F^s_{c}(x^s)$ of a source sample $x^s$ for a certain category $c$ by:
\begin{equation}
    F_{c}^{s}(x^s)=\frac{1}{\left|\Lambda_{c}^{s}\right|}
    y_{c}^{s}\otimes
    \left.f_{E}\left(x^{s}\right)\right|_{c},
\end{equation}
where $y_{c}^{s}$ denotes the label map for category $c$,
$\left.f_{E}\left(x^{s}\right)\right|_{c}$ is the networks' output for category $c$,
$\otimes$ denotes element-wise multiplication for extraction of category-exclusive information,
and $\left|\Lambda_{c}^{s}\right|$ is the number of pixels belonging to the specific category.
The final feature vector $F^{s}(x^s)$ of the source image $x^s$ is obtained by first flattening the $F_{c}^{s}(x)$ of each category into a vector followed by connecting the vectors of all categories into a long vector.
Then, we apply the K-means method~\cite{macqueen1967some} to feature vectors of all source images to group them into $K$ clusters, by minimizing the following error:
\begin{equation}
    \sum_{k=1}^{K} \sum_{x \in \mathcal{C}_k}\left\|F^{s}(x^s)-A^s_{k}\right\|_{2}^{2},
\end{equation}
where $\left\|\cdot\right\|_{2}^{2}$ denotes the $L2$ distance, and $A^s_k$ is the centroid of the cluster $\mathcal{C}_k$:
\begin{equation}
    A^{s}_{k}=\frac{1}{\left|\mathcal{C}_{k}\right|} \sum_{x \in \mathcal{C}_{k}} F^{s}(x^s),
\label{eq:centroid}
\end{equation}
where $\left|\mathcal{C}_{k}\right|$ denotes the number of images belonging to $\mathcal{C}_k$.
The centroids $\{A^s_k\}$ are used as the source-domain anchors, against which the target images will be compared for active sample selection. Note that the cluster number $K$ is not the same as the number of segmentation category $C$, and the impact of different $K$ is explored in Section~\ref{sec:impact_of_K}.

\noindent
\textbf{Active Target Sample Selection Against Source Anchors.}
For single-domain AL, uncertainty-based metrics were extensively used to select the samples which are the most difficult to segment~\cite{siddiqui2020viewal}.
For multi-domain AL, however, we argue that the more dissimilar the target samples are to the source-domain, the more complimentary they are to the segmentation network. Here, we measure the dissimilarity by the distance between the target-domain samples and the source-domain anchors
to assess the importance of unlabeled target-domain samples to domain adaptation.
Specifically, we first calculate the per category feature map of a target-domain image $x^t$:
\begin{equation}
    F_{c}^{t}(x^t)=\frac{1}{\left|\Lambda_{c}^{t}\right|}
    \hat{y}_{c}^{t}\otimes
    \left.f_{E}\left(x^{t}\right)\right|_{c},
\end{equation}
where $\hat{y}_{c}^{t}$ is the predicted label map for category $c$,
and $\left|\Lambda_{c}^{t}\right|$ is the number of pixels belonging to the specific category according to $\hat{y}_{c}^{t}$.
Then, we combine $F_{c}^{t}(x^t)$ of all categories to obtain the image-level feature vector $F^{t}(x^t)$.
Eventually, we calculate the $L2$ distances from $F^{t}(x^t)$ to all source-domain anchors, and define the smallest of them as the distance from the target-domain sample to the source domain:
\begin{equation}
    D(x^t)=\min_k \left\|F^{t}(x^t)-A_{k}^{s}\right\|_{2}^{2}.
\label{eq:distance}
\end{equation}
Intuitively, this definition assigns the target-domain sample to the closest anchor of the source domain's, which corresponds to a mode in the multimodal source-domain distribution.
Based on the distance, we can identify the target-domain samples that are far away from the entire source domain and thus are expected to contain target domain specific information.
Therefore, we select them as active samples and annotate them for subsequent training, hoping to learn unique components of the target-domain distribution from these active annotations.

\subsection{Semi-supervised Domain Adaptation}
\noindent
\textbf{Step-1: Injecting Target-domain Specific Knowledge.}
The actively selected and annotated target-domain samples are added to the training process to learn information exclusive to the target domain (Fig. \ref{fig:framework}(b)).
Training data in this step consist of two parts: the labeled source samples $X^s$ and the active target samples $X^t_L$, and the model $f_E$ is fine-tuned with typical cross-entropy based segmentation losses:
\begin{equation}
    \mathcal{L}_{seg}=\mathcal{L}_{CE}\left(x^{s}, y^{s}\right)+\mathcal{L}_{CE}\left(x^{t}_{L}, y^{t}_{L}\right),
\label{eq:supervise_loss}
\end{equation}
where 
the cross-entropy loss $\mathcal{L}_{CE}$ is defined as:
\begin{equation}
    \mathcal{L}_{CE}=-\frac{1}{HW}\sum_{i=1}^{H \times W}\sum_{c=1}^{C}{y}_{i,c}\log\left(p_{i,c}\right),
\end{equation}
where $y_i$ denotes the label for pixel $i$, and 
$p_i$ is the probability predicted by the model $f_C(f_E)$, and $f_C$ is a classifier. 
As experimentally validated (Section~\ref{sec:sample_selection}), our multi-anchor based active sample selecting strategy is superior to previous strategies, and the model gets a steady improvement in performance with the actively selected samples.

\noindent
\textbf{Step-2: Computing Target-domain Anchors and Pseudo Labels.}
To fully utilize the unlabeled target data $X^t_U$, we use the fine-tuned model to compute pseudo labels $\{\hat{y}^t\}$ for unlabeled target-domain samples as well as target-domain anchors $\{A^t_v\}_{v=1}^V$ (Fig. \ref{fig:framework}(c)), where $V$ represents the number of target-domain anchors.
Notably, as the target-domain anchors are a potentially biased estimation of the actual target-domain distribution, it is natural to correct them dynamically. As indicated by Xie et al.~\cite{xie2016unsupervised}, re-clustering at each
epoch could lead to the collapse of the training process due
to jumps in cluster centroids between epochs. Therefore, we treat the target-domain anchors as a memory bank, and employ the exponential moving average (EMA)~\cite{tarvainen2017mean} to progressively update each anchor in a smooth manner:
\begin{equation}
    A_{v}^{t}=\alpha A_{v}^{t}+(1-\alpha) F^{t}(x^t),
\label{eq:EMA}
\end{equation}
where $\alpha$ is set to 0.999 following~\cite{tarvainen2017mean}, and $F^{t}(x^t)$ is utilized to update the closest anchor. 
With both $\{\hat{y}^t\}$ and $\{A^t_v\}$ computed, we proceed to the next step for semi-supervised domain adaptation.


\noindent
\textbf{Step-3: Semi-supervised Adaptation.}
Lastly, we combine the source data $X^s$, labeled target samples $X^t_L$, and unlabeled target samples $X^t_U$ for a semi-supervised training (\emph{i.e.}, a further fine-tuning of $f_E$) for domain adaptation (Fig. \ref{fig:framework}(d)).
Notably, we propose a novel soft alignment loss to explicitly close the gap between the sample features and anchors in the target domain:

\newcommand{\slfrac}[2]{\left.#1\middle/#2\right.} 
\begin{equation}
    \mathcal{L}_{dis}^{t}= {V}\Big/{{\sum}_{v=1}^{V}\frac{1}{\left\|F^{t}(x^t)-A_{v}^{t}\right\|_{2}^{2}}}.
\label{eq:dis_loss}
\end{equation}
Intuitively, by minimizing the soft alignment loss, features of the target-domain samples output by the model are drawn towards the target-domain anchors,
encouraging a more faithful learning of the underlying target-domain distribution represented by these anchors.
Besides, to make a full use of $X^t_U$, we exploit the pseudo labels ${\hat{y}^{t}}$ to provide further supervision:
\begin{equation}
        \mathcal{L}_{pseudo}=\mathcal{L}_{CE}\left(x^{t}_{U}, \hat{y}^{t}\right).
\label{eq:pseudo_loss}
\end{equation}
Thus, the overall loss function for the semi-supervised learning can be formulated as:
\begin{equation}
        \mathcal{L}_{semi}=\mathcal{L}_{seg}+\mathcal{L}_{dis}^{t}+\mathcal{L}_{pseudo}.
\label{eq:overall_loss}
\end{equation}
The entire training pipeline is summarized in Algorithm \ref{alg:DCCS}.

\begin{algorithm}[t]
	\caption{Multi-anchor Active Domain Adaptation (MADA)}
	\label{alg:DCCS}
	\begin{algorithmic}[1]\small
	    \renewcommand{\algorithmicrequire}{\textbf{Notation:}}
		\REQUIRE  Source-domain set $\left\{\left(x^{s}, y^{s}\right)\right\}$, selected active sample set $\left\{\left(x^{t}_{L}, y^{t}_{L}\right)\right\}$, and unlabeled target-domain set $\left\{x^{t}_{U}\right\}$. Encoder $f_E$, feature vector set of the source domain $\left\{F^{s}(x^s)\right\}$ and feature vector set of the target domain $\left\{F^{t}(x^t)\right\}$. Number of iterations $N$.\\
		\renewcommand{\algorithmicensure}{\textbf{Stage 1:}}
		\ENSURE
		\STATE Warm-up $f_E$ with adversarial training~\cite{tsai2018learning} to obtain $\left\{F^{s}(x^s)\right\}$.\\
		\STATE Apply K-means on $\left\{F^{s}(x^s)\right\}$ to group the source-domain samples into $K$ clusters; \\
		\STATE Compute the centroid $A_k^s$ of the clusters (Eq.~(\ref{eq:centroid})) to serve as the source-domain anchors;  \\
		\STATE Calculate the distance from each target-domain sample to $\left\{A_k^s\right\}$ (Eq.~(\ref{eq:distance})); \\
		\STATE Select 5\% target-domain samples with the smallest distances as active samples for annotation, getting set $\left\{\left(x^{t}_{L}, y^{t}_{L}\right)\right\}$. \\
		\renewcommand{\algorithmicensure}{\textbf{Stage 2:}}
		\ENSURE
		\STATE Fine-tune $f_E$ with both $\left\{\left(x^{s}, y^{s}\right)\right\}$ and $\left\{\left(x^{t}_{L}, y^{t}_{L}\right)\right\}$ by minimizing $\mathcal{L}_{seg}$  (Eq.~(\ref{eq:supervise_loss})), and obtain $\left\{F^{t}(x^t)\right\}$;\\
		\STATE Initialize $A_v^t$ with K-means clustering on $\left\{F^{t}(x^t)\right\}$;\\
		\STATE \textbf{for} $i=1,...,N$ \textbf{do}\\
		\STATE \quad Calculate $\mathcal{L}_{seg}$ (Eq.~(\ref{eq:supervise_loss})) with $\left\{\left(x^{s}, y^{s}\right)\right\}$ and  $\left\{\left(x^{t}_{L}, y^{t}_{L}\right)\right\}$;\\
		\STATE \quad Calculate $\mathcal{L}_{dis}^t$ (Eq.~(\ref{eq:dis_loss})) with $\left\{x^{t}\right\}$ and $\mathcal{L}_{pseudo}$ (Eq.~(\ref{eq:pseudo_loss})) with $\left\{x^{t}_U\right\}$;\\
		\STATE \quad Update $f_E$ by gradient descending $\nabla(\mathcal{L}_{seg}+\mathcal{L}_{dis}^t+\mathcal{L}_{pseudo})$ (Eq.~(\ref{eq:overall_loss}));\\
		\STATE \quad Update $A_v^t$ with EMA (Eq.~(\ref{eq:EMA}));\\
		\STATE \textbf{end for}
	\end{algorithmic}
\end{algorithm}
\vspace{-0.3cm}

\begin{table*}
    \caption{Comparison with other DA methods on the GTA5 to Cityscapes adaptation task. Best results are shown in \textbf{bold}.}
	\renewcommand{\arraystretch}{1.1}
	\small
	\begin{center}
	    
	\scalebox{1.0}{
	\begin{tabular}{C{16.65cm}}
	\toprule[1pt]
	     GTA5 $\rightarrow$ Cityscapses \\
	\end{tabular}}
	\scalebox{0.9}{
    \begin{tabular}{b{2.0cm} b{0.4cm}b{0.4cm}b{0.4cm}b{0.4cm}b{0.4cm}b{0.4cm}b{0.4cm}b{0.4cm}b{0.4cm}b{0.4cm}b{0.4cm}b{0.4cm}b{0.4cm}b{0.4cm}b{0.4cm}b{0.4cm}b{0.4cm}b{0.4cm}b{0.2cm} b{0.7cm}}
    \toprule[0.5pt]
         Method &\rotatebox{90}{road} &\rotatebox{90}{sidewalk} &\rotatebox{90}{building} &\rotatebox{90}{wall} &\rotatebox{90}{fence} &\rotatebox{90}{pole} &\rotatebox{90}{light} &\rotatebox{90}{sign} &\rotatebox{90}{veg} &\rotatebox{90}{terrain} &\rotatebox{90}{sky} &\rotatebox{90}{person} &\rotatebox{90}{rider} &\rotatebox{90}{car} &\rotatebox{90}{truck} &\rotatebox{90}{bus} &\rotatebox{90}{train} &\rotatebox{90}{mbike} &\rotatebox{90}{bicycle} &mIoU  \\
         \hline
    \end{tabular}}
    \scalebox{0.9}{
    \begin{tabular}{b{1.9cm} b{0.4cm}b{0.4cm}b{0.4cm}b{0.4cm}b{0.4cm}b{0.4cm}b{0.4cm}b{0.4cm}b{0.4cm}b{0.4cm}b{0.4cm}b{0.4cm}b{0.4cm}b{0.4cm}b{0.4cm}b{0.4cm}b{0.4cm}b{0.4cm}b{0.4cm} b{0.6cm}}
         AdaptSeg~\cite{tsai2018learning} &86.5 &25.9 &79.8 &22.1 &20.0 &23.6 &33.1 &21.8 &81.8 &25.9 &75.9 &57.3 &26.2 &76.3 &29.8 &32.1 &7.2 &29.5 &32.5 &41.4 \\
         CLAN~\cite{luo2019taking} &87.0 &27.1 &79.6 &27.3 &23.3 &28.3 &35.5 &24.2 &83.6 &27.4 &74.2 &58.6 &28.0 &76.2 &33.1 &36.7 &6.7 &31.9 &31.4 &43.2 \\
         AdvEnt~\cite{vu2019advent} &89.4 &33.1 &81.0 &26.6 &26.8 &27.2 &33.5 &24.7 &83.9 &36.7 &78.8 &58.7 &30.5 &84.8 &38.5 &44.5 &1.7 &31.6 &32.4 &45.5 \\
         BDL~\cite{li2019bidirectional} &91.0 &44.7 &84.2 &34.6 &27.6 &30.2 &36.0 &36.0 &85.0 &43.6 &83.0 &58.6 &31.6 &83.3 &35.3 &49.7 &3.3 &28.8 &35.6 &48.5 \\
         CAG~\cite{zhang2019category} &90.4 &51.6 &83.8 &34.2 &27.8 &38.4 &25.3 &48.4 &85.4 &38.2 &78.1 &58.6 &34.6 &84.7 &21.9 &42.7 &41.1 &29.3 &37.2 &50.2 \\
         \hline
         AADA~\cite{su2020active} &92.2 &59.9 &87.3 &36.4 &45.7 &\textbf{46.1} &50.6 &\textbf{59.5} &88.3 &44.0 &90.2 &69.7 &38.2 &90.0 &55.3 &45.1 &32.0 &32.6 &62.9 &59.3 \\
         MADA (Ours) &\textbf{95.1} &\textbf{69.8} &\textbf{88.5} &\textbf{43.3} &\textbf{48.7} &45.7 &\textbf{53.3} &59.2 &\textbf{89.1} &\textbf{46.7} &\textbf{91.5} &\textbf{73.9} &\textbf{50.1} &\textbf{91.2} &\textbf{60.6} &\textbf{56.9} &\textbf{48.4} &\textbf{51.6} &\textbf{68.7} &\textbf{64.9} \\
    \bottomrule[1pt]
    \end{tabular}}
	\end{center}

    \vspace{-0.3cm}
    \label{table:comparison-GTA5}
\end{table*}

\begin{table*}
    \caption{Comparison with other DA methods on the SYNTHIA to Cityscapes adaptation task. Best results are shown in \textbf{bold}.}
	\renewcommand{\arraystretch}{1.0}
	\small
	\begin{center}
	\scalebox{1.0}{
	\begin{tabular}{C{16.1cm}}
	\toprule[1pt]
	     SYNTHIA $\rightarrow$ Cityscapses \\
	\end{tabular}}
	\scalebox{0.95}{
    \begin{tabular}{b{2.0cm} b{0.4cm}b{0.4cm}b{0.4cm}b{0.4cm}b{0.4cm}b{0.4cm}b{0.4cm}b{0.4cm}b{0.4cm}b{0.4cm}b{0.4cm}b{0.4cm}b{0.4cm}b{0.4cm}b{0.4cm}b{0.2cm} b{0.6cm} b{0.6cm}}
    \toprule[0.5pt]
         Method &\rotatebox{90}{road} &\rotatebox{90}{sidewalk} &\rotatebox{90}{building} &\rotatebox{90}{wall} &\rotatebox{90}{fence} &\rotatebox{90}{pole} &\rotatebox{90}{light} &\rotatebox{90}{sign} &\rotatebox{90}{veg}  &\rotatebox{90}{sky} &\rotatebox{90}{person} &\rotatebox{90}{rider} &\rotatebox{90}{car}  &\rotatebox{90}{bus}  &\rotatebox{90}{mbike} &\rotatebox{90}{bicycle} &mIoU &mIoU*  \\
    \end{tabular}}
    \scalebox{0.95}{
    \begin{tabular}{b{1.9cm} C{0.4cm}C{0.4cm}C{0.4cm}C{0.4cm}C{0.4cm}C{0.4cm}C{0.4cm}C{0.4cm}C{0.4cm}C{0.4cm}C{0.4cm}C{0.4cm}C{0.4cm}C{0.4cm}C{0.4cm}C{0.4cm} C{0.6cm} C{0.5cm}}
         \hline
         AdaptSeg~\cite{tsai2018learning} &79.2 &37.2 &78.8 &- &- &- &9.9 &10.5 &78.2 &80.5 &53.5 &19.6 &67.0 &29.5 &21.6 &31.3 &- &45.9 \\
         CLAN~\cite{luo2019taking} &81.3 &37.0 &80.1 &- &- &- &16.1 &13.7 &78.2 &81.5 &53.4 &21.2 &73.0 &32.9 &22.6 &30.7 &- &47.8 \\
         AdvEnt~\cite{vu2019advent} &85.6 &42.2 &79.7 &8.7 &0.4 &25.9 &5.4 &8.1 &80.4 &84.1 &57.9 &23.8 &73.3 &36.4 &14.2 &33.0 &41.2 &- \\
         BDL~\cite{li2019bidirectional} &86.0 &46.7 &80.3 &- &- &- &14.1 &11.6 &79.2 &81.3 &54.1 &27.9 &73.7 &42.2 &25.7 &45.3 &- &51.4 \\
         CAG~\cite{zhang2019category} &84.7 &40.8 &81.7 &7.8 &0.0 &35.1 &13.3 &22.7 &84.5 &77.6 &64.2 &27.8 &80.9 &19.7 &22.7 &48.3 &44.5 &50.9\\
         \hline
         AADA~\cite{su2020active} &91.3 &57.6 &86.9 &37.6 &\textbf{48.3} &45.0 &50.4 &58.5 &88.2 &90.3 &69.4 &37.9 &89.9 &44.5 &32.8 &62.5 &61.9 &66.2 \\
         MADA (Ours) &\textbf{96.5} &\textbf{74.6} &\textbf{88.8} &\textbf{45.9} &43.8 &\textbf{46.7} &\textbf{52.4} &\textbf{60.5} &\textbf{89.7} &\textbf{92.2} &\textbf{74.1} &\textbf{51.2} &\textbf{90.9} &\textbf{60.3} &\textbf{52.4} &\textbf{69.4} &\textbf{68.1} &\textbf{73.3} \\
    \bottomrule[1pt]
    \end{tabular}}
	\end{center}

    \vspace{-0.5cm}
    \label{table:comparison-SYN}
\end{table*}

\section{Experiments}

\subsection{Datasets}
To demonstrate the superiority of our proposed method, two challenging \textit{synthia-2-real} adaptation tasks, \textit{i.e.},  GTA5~\cite{richter2016playing} $\rightarrow$ Cityscapes~\cite{cordts2016cityscapes} and SYNTHIA~\cite{ros2016synthia} $\rightarrow$ Cityscapes are applied for evaluation. To be specific:

\begin{itemize}
\item GTA5 $\rightarrow$ Cityscapes: The GTA5 dataset consists of 24,966 synthetic images with 19-class segmentation, which is consistent with the Cityscapses dataset.
\item SYNTHIA $\rightarrow$ Cityscapes: Following the previous study~\cite{li2019bidirectional}, the SYNTHIA-RAND-CITYSCAPES set with 9,400 synthetic images containing 16-class segmentation is utilized for training.
\end{itemize}

In both sets, Cityscapes serves as the target domain, with 2,975 images for training and 500 images for evaluation. The segmentation performance is measured with the mean-Intersection-over-Union (mIoU)~\cite{everingham2015pascal} metric.

\subsection{Implementation Details}
We employ the DeepLab v3+~\cite{chen2018encoder} as the feature extractor $f_{E}$, which is composed of the backbone ResNet-101~\cite{he2016deep} pretrained on ImageNet~\cite{deng2009imagenet} and the Atrous Spatial Pyramid Pooling (ASPP) module. The classifier $f_{C}$ is a typical convolutional layer with $C$ channels and 1 $\times$ 1 kernel size to transform the latent representation to semantic segmentation.
During the warm-up, the discriminator $f_{D}$ consists of 5 convolutional layers of kernel size 3 $\times$ 3 and stride 2 with numbers of filters set to $\{64, 128, 256, 512, 1\}$. The first three convolutional layers are followed with a Rectified Linear Unit (ReLU) layer, while the fourth one is followed by a leaky ReLU~\cite{maas2013rectifier} parameterized by 0.2. The proposed method is implemented on PyTorch with a TITAN Tesla V100 GPU. The input images are randomly resized with a ratio in $[0.5, 1.5]$
and then cropped to 896 $\times$ 512 pixels. 

For warm-up, we train the model for 20 epochs in an adversarial manner with a cross entropy loss and an adversarial loss weighted by $0.01$. For fine-tuning in the second stage, we use the SGD optimizer to train our model for 50 epochs. The learning rate is initially set to $2.5 \times 10^{-4}$ and decayed by poly learning rate policy with a power of 0.9.

Except for the comparison study in Section~\ref{sec:impact_of_sampleno}, we select 5\% target-domain samples as active samples for all experiments , which takes little annotation workload but brings large performance gain.

\begin{figure*}[ht]
	\centering
	\includegraphics[width=1.8\columnwidth]{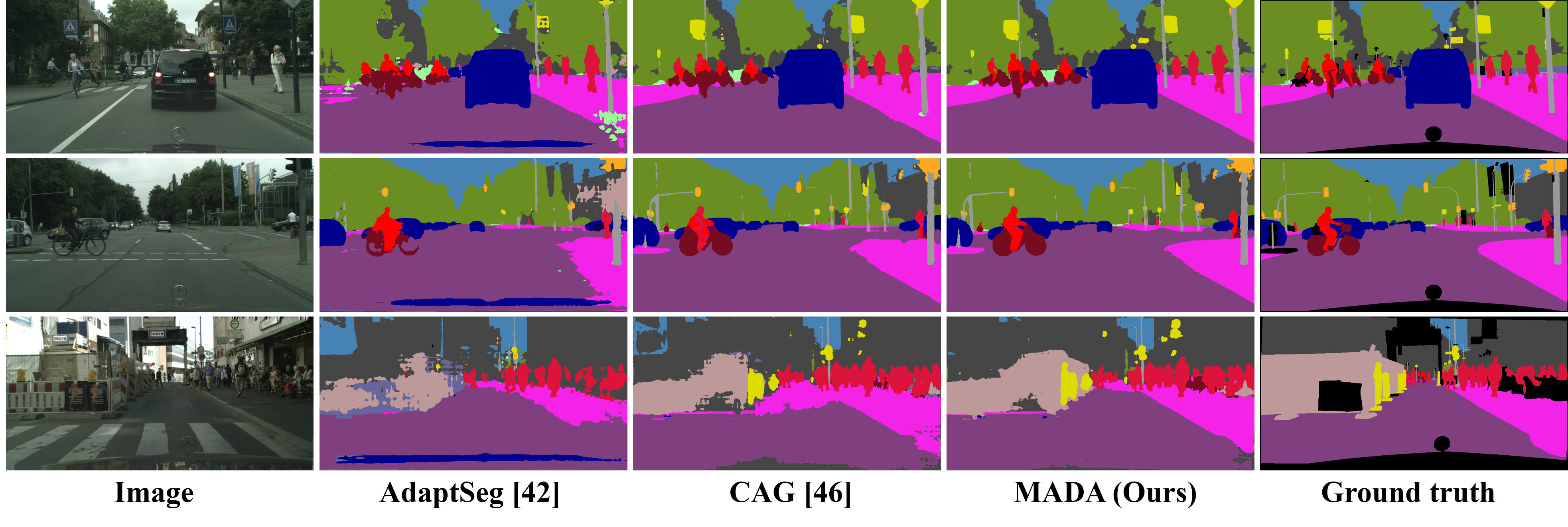}
	\caption{Qualitative results of DA segmentation for GTA5 $\rightarrow$ Cityscapes. For each image, we show the results of the typical adversarial method~\cite{tsai2018learning}, state-of-the-art UDA method~\cite{zhang2019category} and our proposed MADA, respectively. The black region in ``Ground truth'' is excluded from evaluation because it does not belong to any of the 19 classes.}
	\label{fig:visualizatoin}
	\vspace{-0.3cm}
\end{figure*}

\subsection{Main Results}
\label{sec:comparison}
As presented in Table~\ref{table:comparison-GTA5} and Table \ref{table:comparison-SYN}, the proposed framework is compared with five UDA methods~\cite{tsai2018learning,luo2019taking,vu2019advent,li2019bidirectional,zhang2019category} and an active DA approach~\cite{su2020active}. As expected, we observe substantial improvements over the UDA methods, suggesting that with carefully selected active samples, little manual annotation workload can lead to large performance gains. In addition, the proposed method outperforms another active DA method, \textit{i.e.}, AADA, by a large margin (5.6\% mIOU), demonstrating the effectiveness of the proposed multi-anchor strategy. The visualization of three example images, which are the same as those in Fig.~\ref{fig:cluster_visual_1}, is displayed in Fig.~\ref{fig:visualizatoin} for qualitative comparison. We can observe that by alleviating the distortion of target features, fewer segmentation errors as well as more precise boundaries can be obtained with the proposed MADA method.

\subsection{Ablation Study}
\label{sec:ablation}
To verify the effectiveness of each component, we perform an ablation study with the following variants: $\mathbf{M}^{(0)}$: the baseline adversarial learning method \cite{tsai2018learning} without any active annotation; $\mathbf{M}^{(1)}$: extending $\mathbf{M}^{(0)}$ by additionally introducing the active samples with cross entropy loss for training; $\mathbf{M}^{(2)}$: extending $\mathbf{M}^{(1)}$ by adding the proposed multi-anchor soft alignment loss on target samples for optimization; $\mathbf{M}^{(3)}$: extending $\mathbf{M}^{(2)}$ by progressively updating the target anchors with EMA; $\mathbf{M}^{(4)}$: adding the pseudo label loss for optimization in addition to $\mathbf{M}^{(3)}$; 
$\mathbf{M}^{(u)}$: performing fully-supervised segmentation with the annotation of both the source and target datasets as the upper bound. As shown in Table~\ref{table:ablation_study}, the consistent and notable improvements from $\mathbf{M}^{(0)}$ to $\mathbf{M}^{(4)}$ on two public datasets demonstrate the effectiveness of each strategy.
Furthermore, MADA with only 5\% of the target-domain samples actively annotated achieves a comparable performance with that of the upper bound,
suggesting that the proposed framework can select complimentary samples to effectively close the gap between UDA and full supervision.

The visualization of the feature distribution with/without active learning is presented in Fig~\ref{fig:cluster_visual_1}. With the proposed MADA framework, the target-specific information can be maintained as its original multimodal distribution. 




\begin{table}
    \caption{Ablation study. G $\rightarrow$ C denotes the GTA5 $\rightarrow$ Cityscapes scenario and S $\rightarrow$ C denotes the SYNTHIA $\rightarrow$ Cityscapes scenario.}
    \begin{center}
	\renewcommand{\arraystretch}{1.0}
	\small
	\scalebox{0.9}{
    \begin{tabular}{C{1.4cm}C{0.5cm}C{0.5cm}C{0.5cm}C{0.5cm}C{1.1cm}C{1.1cm}}
    \toprule[1pt]
    & & & & &G $\rightarrow$ C &S $\rightarrow$ C \\
    \toprule[0.5pt]
         Method &A &B &C &D &mIoU &mIoU  \\
         \hline
         $\mathbf{M}^{(0)}$ & & & & &42.5 &42.9\\
         \hline
         $\mathbf{M}^{(1)}$ &\checkmark & & & &61.6 &65.0\\
         \hline
         $\mathbf{M}^{(2)}$ &\checkmark &\checkmark & & &63.2 &66.6\\
        $\mathbf{M}^{(3)}$ &\checkmark &\checkmark &\checkmark & &63.8 &67.6\\
        $\mathbf{M}^{(4)}$ &\checkmark &\checkmark &\checkmark &\checkmark &64.9 &68.1\\
    \hline
    $\mathbf{M}^{(u)}$ & & & & &69.3 &70.8\\
    \hline
    \multicolumn{6}{p{6.6cm}}{A: Training with active samples}\\
    \multicolumn{6}{p{6.6cm}}{B: Soft-anchor alignment loss}\\
    \multicolumn{6}{p{6.6cm}}{C: Updating target anchor with EMA}\\
    \multicolumn{6}{p{6.6cm}}{D: Pseudo training for unlabled target samples}\\
    \bottomrule[1pt]
    \end{tabular}}

    \end{center}
    \label{table:ablation_study}
    \vspace{-0.8cm}
\end{table}


    

\begin{table*}[!htp]
    \caption{Experiments on different active sample selection methods. Best results are shown in \textbf{bold}.}
	\renewcommand{\arraystretch}{1.0}
	\small
	\begin{center}
	    
	\scalebox{1.0}{
	\begin{tabular}{C{16.65cm}}
	\toprule[1pt]
	     GTA5 $\rightarrow$ Cityscapses \\
	\end{tabular}}
	\scalebox{0.9}{
    \begin{tabular}{b{2.0cm} b{0.4cm}b{0.4cm}b{0.4cm}b{0.4cm}b{0.4cm}b{0.4cm}b{0.4cm}b{0.4cm}b{0.4cm}b{0.4cm}b{0.4cm}b{0.4cm}b{0.4cm}b{0.4cm}b{0.4cm}b{0.4cm}b{0.4cm}b{0.4cm}b{0.2cm} b{0.7cm}}
    \toprule[0.5pt]
         Method &\rotatebox{90}{road} &\rotatebox{90}{sidewalk} &\rotatebox{90}{building} &\rotatebox{90}{wall} &\rotatebox{90}{fence} &\rotatebox{90}{pole} &\rotatebox{90}{light} &\rotatebox{90}{sign} &\rotatebox{90}{veg} &\rotatebox{90}{terrain} &\rotatebox{90}{sky} &\rotatebox{90}{person} &\rotatebox{90}{rider} &\rotatebox{90}{car} &\rotatebox{90}{truck} &\rotatebox{90}{bus} &\rotatebox{90}{train} &\rotatebox{90}{mbike} &\rotatebox{90}{bicycle} &mIoU  \\
         \hline
    \end{tabular}}
    \scalebox{0.9}{
    \begin{tabular}{b{1.9cm} b{0.4cm}b{0.4cm}b{0.4cm}b{0.4cm}b{0.4cm}b{0.4cm}b{0.4cm}b{0.4cm}b{0.4cm}b{0.4cm}b{0.4cm}b{0.4cm}b{0.4cm}b{0.4cm}b{0.4cm}b{0.4cm}b{0.4cm}b{0.4cm}b{0.4cm} b{0.6cm}}
         Random  &92.8 &64.5 &85.8 &38.0 &34.8 &43.7 &50.1 &56.9 &87.9 &40.4 &87.7 &69.0 &30.8 &89.4 &51.1 &43.8 &21.7 &29.9 &59.4 &56.7 \\
         Entropy~\cite{vu2019advent} &93.9 &65.4 &87.7 &42.2 &48.4 &46.7 &47.3 &57.0 &88.5 &44.3 &90.4 &70.8 &32.8 &90.0 &53.8 &49.9 &30.0 &41.1 &63.6 &60.2 \\
         Adversarial~\cite{tsai2018learning} &91.8 &59.2 &87.5 &37.8 &45.2 &45.5 &51.5 &56.9 &88.5 &43.0 &90.3 &69.0 &37.1 &89.9 &54.5 &46.1 &35.9 &28.1 &61.3 &58.9 \\
         AADA~\cite{su2020active} &92.2 &59.9 &87.3 &36.4 &45.7 &\textbf{46.1} &\textbf{50.6} &\textbf{59.5} &\textbf{88.3} &\textbf{44.0} &\textbf{90.2} &69.7 &38.2 &\textbf{90.0} &\textbf{55.3} &45.1 &32.0 &32.6 &62.9 &59.3 \\
         Proposed &\textbf{92.4} &\textbf{61.4} &\textbf{87.4} &\textbf{39.5} &\textbf{45.9} &45.2 &\textbf{50.6} &57.5 &87.8 &42.4 &89.2 &\textbf{72.7} &\textbf{44.9} &\textbf{90.0} &54.7 &\textbf{50.5} &\textbf{43.4} &\textbf{47.8} &\textbf{66.9} &\textbf{61.6} \\
    \bottomrule[1pt]
    \end{tabular}}
	\end{center}
    \label{table:selection}
    \vspace{-0.5cm}
\end{table*}

\subsection{Comparison of Sample Selection Methods}
\label{sec:sample_selection}
The performance of active learning depends heavily on the sample selection methods. On Table~\ref{table:selection}, we compare the proposed anchor-based method with the following popular sample selection approaches on the GTA5 to Cityscapes adaptation task. 

\noindent
\textbf{Random Selection.} Samples are randomly selected with equal probability from the target domain.

\noindent
\textbf{Entropy-based Uncertainty Method.} The AdvEnt~\cite{vu2019advent} is applied to obtain the prediction map entropy of each sample in the target domain and the ones with top $5\%$ entropy are chosen for manual annotation:
\begin{equation}
    {E}_{{ent}}=\frac{-1}{\log (C)} \sum_{c=1}^{C}\sum_{i=1}^{H \times W} p_{i,c}^{t} \log (p_{i,c}^{t}).
\end{equation}

\noindent
\textbf{Adversarial-based Diversity Method.} With the discriminator $f_D$ trained in the warm-up stage as~\cite{tsai2018learning}, we select the samples with least predicted probabilities, \textit{i.e.}, the ones that are most distinguishable from the source domain:
\begin{equation}
    {E}_{adv}=\frac{1-f_D(f_E(x^t)}{f_D(f_E(x^t))}.
\end{equation}

\noindent
\textbf{AADA Method.} In addition to the discriminator-based diversity, the AADA~\cite{su2020active} method also takes the certainty of prediction into consideration: 
\begin{equation}
    E_{AADA}=E_{ent}E_{adv}.
\end{equation}


Note that for a fair comparison, all the comparison experiments are subject to the same experimental setup. The same percentage of active samples, $5\%$, are selected, while no unlabeled samples are used for optimization. We can observe that the proposed multi-anchor strategy delivers the best segmentation performance in mIoU, suggesting that better active samples are selected by our proposed strategy.

\subsection{Impact of the Number of Anchors}
\label{sec:impact_of_K}
\begin{figure}[!htp]
	\centering
	\includegraphics[width=0.90\columnwidth]{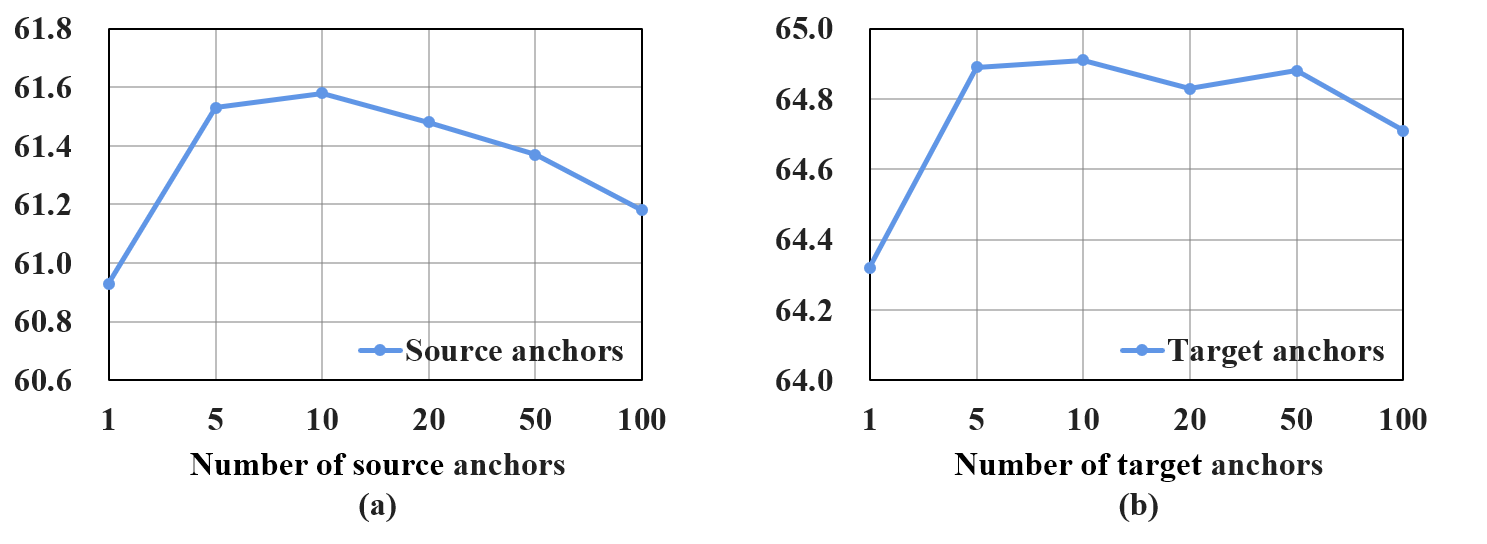}
	\caption{Experiments on different number of anchors for the source domain (a) and target domain (b).}
	\label{fig:n_of_anchors}
	\vspace{-0.5cm}
\end{figure}

We evaluate the impact of different anchor numbers on modeling the source and target domains with the
GTA5 to Cityscapes adaptation task, where the number of anchors varies from 1 to 100 in both domains. As shown in Fig.~\ref{fig:n_of_anchors}, for both domains, using multiple anchors was consistently better than using a single centroid, and using 5--10 anchors stably yielded superior performance.
This might be because there are only limited types of scenarios in these datasets, and a few anchors are sufficient to represent their distributions.
We therefore use 10 clusters considering the top performance in both domains.

\subsection{Impact of the Number of Active Samples}
\label{sec:impact_of_sampleno}
In order to verify the stability of our proposed method, comparative experiments for different percentages of active samples are conducted. As shown in Table~\ref{table:partion}, as the percentage of samples increases from 1\% to 20\%, the mIoU increases steadily from 56.7\% to 64.1\%. We also introduce the upper bound by optimizing with all target labels, the narrow gap of 7.7\% in mIoU  between using only 5\% of target-domain data for AL and the upper bound demonstrates that the proposed method can effectively exploit the information from active samples.

\begin{table}[!htp]
    \centering
    \caption{Experiments on different number of active samples.}
    \small
	\renewcommand{\arraystretch}{0.9}
	\scalebox{0.85}{
	\begin{tabular}{C{1.5cm} C{0.7cm}C{0.7cm}C{0.7cm}C{0.7cm}C{0.7cm}C{0.7cm}}
    \toprule[1pt]
    \multicolumn{7}{c}{GTA5 $\rightarrow$ Cityscapse}\\
    \toprule[0.5pt]
    Percentage &1\% &2\% &5\% &10\% &20\% &100\% \\
    \toprule[0.5pt]
    mIoU &56.7 &59.1 &61.6 &62.7 &64.1 &69.3 \\
    \toprule[0.5pt]
    mIoU Gap &$-$12.6 &$-$10.2 &$-$7.7 &$-$6.6 &$-$5.2 &-\\
    \bottomrule[1pt]
    \end{tabular}}
    \label{table:partion}
    \vspace{-0.5cm}
\end{table}

\section{Conclusion}
In this paper, we proposed the Multi-anchor Active Domain Adaptation (MADA) framework, for distortion-free source-to-target domain adaptation of segmentation models at minimal annotation cost.
MADA introduced anchor-based active sample selection into DA, for selection of limited target-domain samples that were most complementary to the source-domain distribution and meanwhile unique to the target-domain distribution.
Adding active annotation of these selected target-domain samples for training can effectively prevent distortion of the target-domain distribution that could otherwise happen in typical UDA methods.
Different from previous works which assumed unimodal distributions for both the source and target domains, MADA proposed to use multiple anchors to realize multimodal distributions for both domains.
On top of that, MADA further proposed a multi-anchor soft-alignment loss to explicitly push the target-domain features towards these anchors, for full utilization of the unlabeled target-domain samples.
Experimental results on two public benchmark datasets demonstrated the effectiveness of (i) introducing AL into DA, (ii) multiple anchors versus a single centroid,  and (iii) adding the soft-alignment loss, as well as the superior performance of MADA towards existing state-of-the-art UDA and active DA methods.

\clearpage
{\small
\bibliographystyle{ieee_fullname}
\bibliography{ref}
}

\clearpage
\section*{Appendix}
\subsection*{A. Imapact of the warm-up model}
The impact of warm-up models of various capability (as reflected by the UDA mIoU), including `inferior', `standard' (used in our paper), and `superior',
on the active learning (AL) performance is charted in Table~\ref{table:warmup}. The results demonstrate that the AL performance is unaffected, suggesting the robustness of our proposed AL criterion against mistakes of the warm-up model.

\begin{table}[!htp]
    \centering
    \small
    \caption{Impact of the warm-up model (GTA5 $\rightarrow$ Cityscapse).}
	\renewcommand{\arraystretch}{0.85}
	\scalebox{0.9}{
	\begin{tabular}{C{2.1cm} C{1.8cm}C{1.8cm}C{1.8cm}}
    \toprule[1pt]
    Warm-up model &Inferior & Standard & Superior \\
    \toprule[0.5pt]
    UDA mIoU &40.65 &42.53 &44.00  \\
    \toprule[0.5pt]
    AL mIoU &61.36 &61.58 &61.50  \\
    
    \bottomrule[1pt]
    \end{tabular}}
    \label{table:warmup}
    
\end{table}

\subsection*{B. Impact of clustering methods}
As shown in Table~\ref{table:clustering}, K-means (61.58 mIoU) outperforms spectral clustering (61.02 mIoU) 
when integrated in MADA. 
In addition, K-means incurs less computational time in our exploratory experiments.
Therefore, we use K-means for our framework.

\begin{table}[!htp]
    \centering
    \small
    \caption{Impact of the clustering methods (GTA5 $\rightarrow$ Cityscapse).}
	\renewcommand{\arraystretch}{0.85}
	\scalebox{0.9}{
	\begin{tabular}{C{3.0cm} C{1.8cm}}
    \toprule[1pt]
    Clustering method &mIoU \\
    \toprule[0.5pt]
    Spectral clustring &61.02 \\
    \toprule[0.5pt]
    K-means &61.58 \\
    
    \bottomrule[1pt]
    \end{tabular}}
    \label{table:clustering}
    
\end{table}

\end{document}